\documentclass[letterpaper, 10 pt, conference]{ieeeconf}  

\IEEEoverridecommandlockouts                              

\usepackage{graphics} 
\usepackage{epsfig} 
\usepackage{mathptmx} 
\usepackage{times} 
\usepackage{amsmath} 
\usepackage{amssymb}  
\usepackage{hyperref}
\usepackage{multirow}

\title{\LARGE \bf
L2RDaS: Synthesizing 4D Radar Tensors for Model Generalization via Dataset Expansion
}

\author{Woo-Jin Jung, Dong-Hee Paek, and Seung-Hyun~Kong* 
\thanks{This work was supported by the National Research Foundation of Korea(NRF) grant funded by the Korea government(MSIT) (No. 2021R1A2C3008370). \textit{(Corresponding author: Seung-Hyun Kong)}}
\thanks{Woo-Jin Jung, Dong-Hee Paek, and Seung-Hyun Kong are with the CCS Graduate School of Mobility, Korea Advanced Institute of Science and Technology, Daejeon, Korea, 34051. (e-mail: \{woo-jin.jung, donghee.paek, skong\}@kaist.ac.kr)}}

\begin{document}

\maketitle
\thispagestyle{empty}
\pagestyle{empty}

\begin{abstract}

4-dimensional (4D) radar is increasingly adopted in autonomous driving for perception tasks, owing to its robustness under adverse weather conditions. To better utilize the spatial information inherent in 4D radar data, recent deep learning methods have transitioned from using sparse point cloud to 4D radar tensors. However, the scarcity of publicly available 4D radar tensor datasets limits model generalization across diverse driving scenarios. Previous methods addressed this by synthesizing radar data, but the outputs did not fully exploit the spatial information characteristic of 4D radar. To overcome these limitations, we propose LiDAR-to-4D radar data synthesis (L2RDaS), a framework that synthesizes spatially informative 4D radar tensors from LiDAR data available in existing autonomous driving datasets. L2RDaS integrates a modified U-Net architecture to effectively capture spatial information and an object information supplement (OBIS) module to enhance reflection fidelity. This framework enables the synthesis of radar tensors across diverse driving scenarios without additional sensor deployment or data collection. L2RDaS improves model generalization by expanding real datasets with synthetic radar tensors, achieving an average increase of 4.25\% in ${{AP}_{BEV}}$ and 2.87\% in ${{AP}_{3D}}$ across three detection models. Additionally, L2RDaS supports ground-truth augmentation (GT-Aug) by embedding annotated objects into LiDAR data and synthesizing them into radar tensors, resulting in further average increases of 3.75\% in ${{AP}_{BEV}}$ and 4.03\% in ${{AP}_{3D}}$. The implementation will be available at \url{https://github.com/kaist-avelab/K-Radar}.

\end{abstract}


\section{Introduction} \label{sec:introduction}

\begin{figure}[!th]
    \centering
    \includegraphics[width=0.47\textwidth]{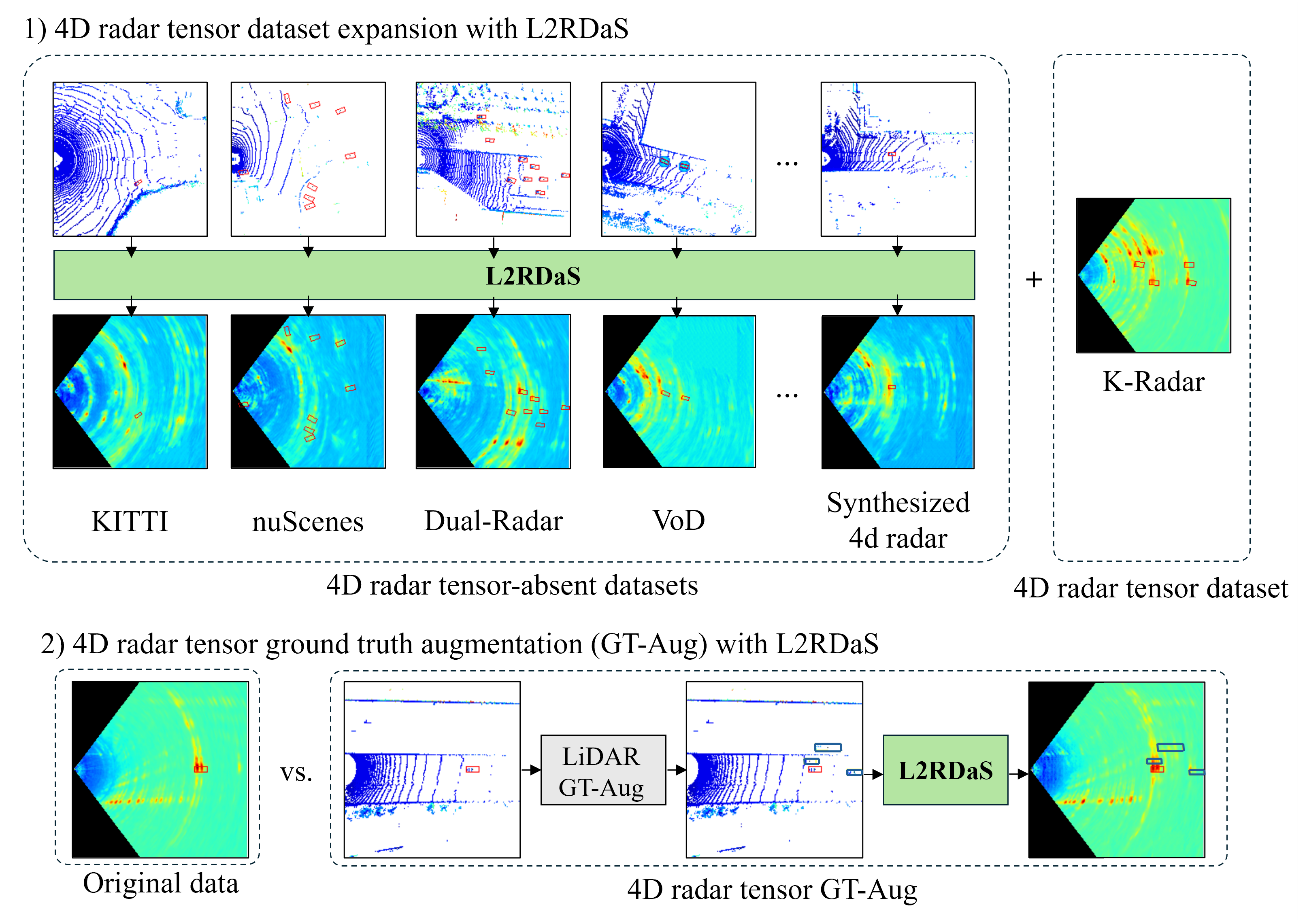} 
    \caption{Application of the L2RDaS. L2RDaS is a framework designed to synthesize 4D radar tensors from LiDAR data. It can be applied to existing autonomous driving datasets that lack 4D radar tensors, enabling dataset expansion without additional sensor deployment or data collection. This allows models to achieve better generalization across diverse driving scenarios.
Furthermore, L2RDaS supports ground-truth augmentation (GT-Aug) by first augmenting the LiDAR data and then synthesizing it into 4D radar tensors. The resulting tensors preserve realistic object and clutter measurements, allowing GT-Aug to be effectively applied to 4D radar tensor datasets as well.}
    \label{fig:overall_L2RDaS}
\end{figure}

Object detection is a fundamental function of autonomous driving, supplying the positions and classes of surrounding traffic participants for downstream planning and control. Reliable operation therefore hinges on robust detection, defined as consistently accurate performance under varying illumination, precipitation, and visibility conditions. Camera- and LiDAR-based detectors, which rely on visible or near-infrared light, suffer severe degradation in adverse weather such as fog, rain, or snow. In contrast, millimeter-wave radar penetrates atmospheric particles and preserves range, Doppler, and azimuth information. Recent hardware advancements have evolved millimeter-wave radar into 4D radar by incorporating elevation antenna arrays, enabling the capture of spatial information (range, azimuth, elevation) along with Doppler \cite{ieee_tutorial}.

Most 4D radar data are provided as point cloud representations, typically extracted from 4D radar tensors using traditional handcrafted methods such as constant false alarm rate (CFAR) filtering \cite{clutter}. These tensors represent dense grids that encode reflected radar power across discretized dimensions of range, Doppler, azimuth, and elevation. Since 4D radar tensors contain both true object measurements and clutter, CFAR suppresses noise by comparing each cell’s power to a local threshold. While effective at removing clutter, this method generally assumes fixed object sizes through predefined hyperparameters such as guard and training cells, making it unsuitable for driving scenarios with diverse objects and distance-dependent size variations \cite{radelft}. As a result, point cloud obtained via CFAR tend to be sparse, leading to loss of detailed spatial information necessary for accurately representing object shapes and boundaries \cite{rpfa_net}, which in turn results in imprecise geometry and complicates sensor fusion in autonomous driving \cite{dpft}.

Due to the limitations of point cloud representation discussed earlier, there is a growing effort to leverage 4D radar tensors directly for object detection \cite{KRadar, rtnh+, radarocc, CARB, centerradarnet, dpft, echoesfusion}. However, only five public datasets currently provide access to 4D radar tensors \cite{Radial, KRadar, SCORP, RATRON, radelft}, in contrast to over 30 publicly available LiDAR datasets \cite{lidardataset}. This scarcity of publicly available 4D radar tensor datasets poses a significant challenge for training models that generalize across diverse driving scenarios. Insufficient diversity in training data can hinder the deployment of radar-based perception in autonomous driving. The limited availability of 4D radar tensors primarily stems from the recent commercialization of 4D radar sensors (around 2022) and manufacturers’ reluctance to release tensor data due to proprietary concerns \cite{ieee_tutorial}.

To improve model generalization across diverse driving scenarios, researchers have explored synthesizing 4D radar tensors for training using existing autonomous driving datasets or simulation environments \cite{l2r, l2rtranslation_voxel, gan_radar_synth, radsimreal, raids, SHENRON, automotiveRadar_estimation}. However, prior method have primarily focused on synthesizing point cloud representations, which suffer from spatial information loss, or tensor representations that do not exploit the spatial information characteristic of 4D radar.  As demonstrated in the K-Radar study \cite{KRadar}, the RTNH model—a 4D radar tensor-based detection network—significantly outperforms its 2D counterpart (RTN), which uses only planar information, highlighting the importance of fully leveraging the spatial information characteristic of 4D radar.

To overcome the limitations of existing radar synthesis methods, we propose LiDAR-to-4D radar data synthesis (L2RDaS), a framework that synthesizes spatially informative 4D radar tensors (range, azimuth, elevation) from LiDAR data in existing autonomous driving datasets. This enables the synthesizing of radar tensors across diverse driving scenarios without additional sensor deployment or data collection. The L2RDaS framework incorporates a L2RDaS Generator as its core component for synthesizing 4D radar tensors. This L2RDaS Generator modifies a conditional generative adversarial networks (cGAN) based architecture by adopting a tailored U-Net structure to preserve the spatial information characteristic of 4D radar. L2RDaS introduces the Object Information Supplement (OBIS) module to address issues in LiDAR-to-4D radar synthesis, such as low radar resolution and LiDAR point cloud sparsity. By injecting object-level information, OBIS enhances reflection fidelity and spatial consistency in the synthesized radar tensors. The proposed framework synthesizes 4D radar tensors from datasets that do not originally contain 4D radar tensors, such as KITTI, nuScenes, VoD, and Dual-Radar, and expands the real 4D radar tensor dataset K-Radar by incorporating the synthesized data for training. This expansion resulted in average increases of 4.25\% in ${{AP}_{BEV}}$ and 2.87\% in ${{AP}_{3D}}$ across three detection models, demonstrating the effectiveness of the method. Additionally, L2RDaS supports ground-truth augmentation (GT-Aug) by embedding annotated objects into LiDAR data and synthesizing them into 4D radar tensors, which include radar-specific measurements such as realistic clutter. This demonstrates that L2RDaS can effectively augment 4D radar tensors, resulting in further average increases of 3.75\% in ${{AP}_{BEV}}$ and 4.03\% in ${{AP}_{3D}}$.

Our main contributions are given as follows.
\begin{itemize} 
\item We propose L2RDaS, the first framework that synthesizes spatially informative 4D radar tensors (range, azimuth, elevation) from LiDAR data in existing autonomous driving datasets, enabling the expansion of real radar datasets with synthesized data to improve model generalization across diverse driving scenarios.
\item We introduce the OBIS module, which enhances the realism of synthesized 4D radar tensors by addressing issues such as low radar resolution and LiDAR point cloud sparsity, thereby improving reflection fidelity and spatial consistency.
\item We propose a method that enables GT-Aug for 4D radar tensors by embedding annotated objects into LiDAR and synthesizing them into 4D radar tensors, providing a practical solution for data augmentation.
\end{itemize}

This paper is organized as follows. Section~\ref{sec:related_works} reviews prior studies on radar data synthesis from other modalities and introduces the base model for L2RDaS Generator. Section~\ref{sec:method} provides a detailed explanation of the proposed L2RDaS framework, including the challenges encountered during synthesis and the methods used to address them. Section~\ref{sec:experiment} describes the training process, outlines the experimental setup, and presents both quantitative and qualitative results, including an ablation study. Finally, Section~\ref{sec:conclusion} concludes the study and discusses future research directions.

\section{Related work} \label{sec:related_works}
\subsection{Radar Data Generation from Different Modalities}
Recent studies on radar data synthesis can largely be categorized into two groups: (1) deep learning-based methods that synthesize radar data from other sensor modalities, and (2) physics-based simulation methods that replicate radar signals by modeling electromagnetic wave propagation.

\subsubsection{Deep learning-based Synthesis} 
Several methods synthesize radar tensors—primarily in the form of 2D such as Range-Doppler (RD) or Range-Azimuth (RA) tensors—using deep neural networks. Weston et al.\cite{radar_simulate} synthesize radar tensors from elevation maps via adversarial training and cyclic consistency loss. L2R GAN \cite{l2r} synthesizes 2D radar tensors from LiDAR point cloud using a cGAN composed of an occupancy-grid-mask-guided global generator and a local region generator. Fidelis et al. \cite{gan_radar_synth} propose a GAN-based method that takes object distance and Gaussian noise as inputs to synthesize raw FMCW radar signals, generating a sequence of 16 chirps, which are then processed into RA tensors. Alkanat et al. \cite{automotiveRadar_estimation} predict Gaussian Mixture Model (GMM) parameters from RGB images using a CNN to represent radar point cloud distributions, enabling realistic synthesis via probabilistic sampling. RAIDS \cite{raids} synthesizes RA tensors from RGB images, depth, and semantic maps using a convolutional autoencoder enhanced with channel and spatial attention to improve reflection fidelity and localization. Additionally, LiDAR-to-Radar Synthesis \cite{l2rtranslation_voxel} synthesizes radar point clouds directly from LiDAR point cloud using a voxel-based feature extraction module.

\subsubsection{Physics-based Simulation} 
In contrast to deep learning-based methods, physics-based simulation frameworks simulate radar measurements by explicitly modeling the physical process of radar wave propagation and reflection. RadSimReal \cite{radsimreal} takes 3D simulated environments as input and generates annotated RA radar images via ray tracing and radar signal processing. SHENRON \cite{SHENRON} produces high-resolution radar point cloud from sparse LiDAR point cloud and camera images by estimating RF reflection properties through 3D geometry and material information derived from images.

Despite recent advances, most existing methods that incorporate elevation information typically produce point cloud representations. These are highly sensitive to preprocessing steps and prone to spatial information loss, limiting their suitability for deep learning-based perception. While some methods synthesize radar tensors instead of point cloud, they often rely on partial spatial radar tensors, which are insufficient to exploit the spatial information characteristic of 4D radar.

\subsection{L2RDaS Base Model: Image Translation} \label{sec:related_image_translation}
Image translation focuses on transforming sensor modality or style while preserving spatial structure. Notable methods include pix2pix \cite{pix2pix} and pix2pixHD \cite{pix2pixhd}, which use cGAN to translate input images into corresponding outputs. Pix2pix employs a modified U-Net architecture \cite{unet} as a generator with a patch-based discriminator, enabling tasks such as image synthesis and domain adaptation. Pix2pixHD extends this to high-resolution images using boundary maps, multi-scale generators, and multi-scale discriminators, resulting in improved output fidelity.

These models excel at 2D-to-2D translation while maintaining spatial structure. However, L2RDaS aims to synthesize spatially informative 4D radar tensors from 3D LiDAR point cloud, limiting the direct applicability of conventional image translation frameworks. To address this, we extend the pix2pixHD generator by incorporating architectural modifications tailored for synthesizing 4D radar tensors.

\section{Method} \label{sec:method}
\begin{figure*}[!th]
    \centering
    \includegraphics[width=1\textwidth]{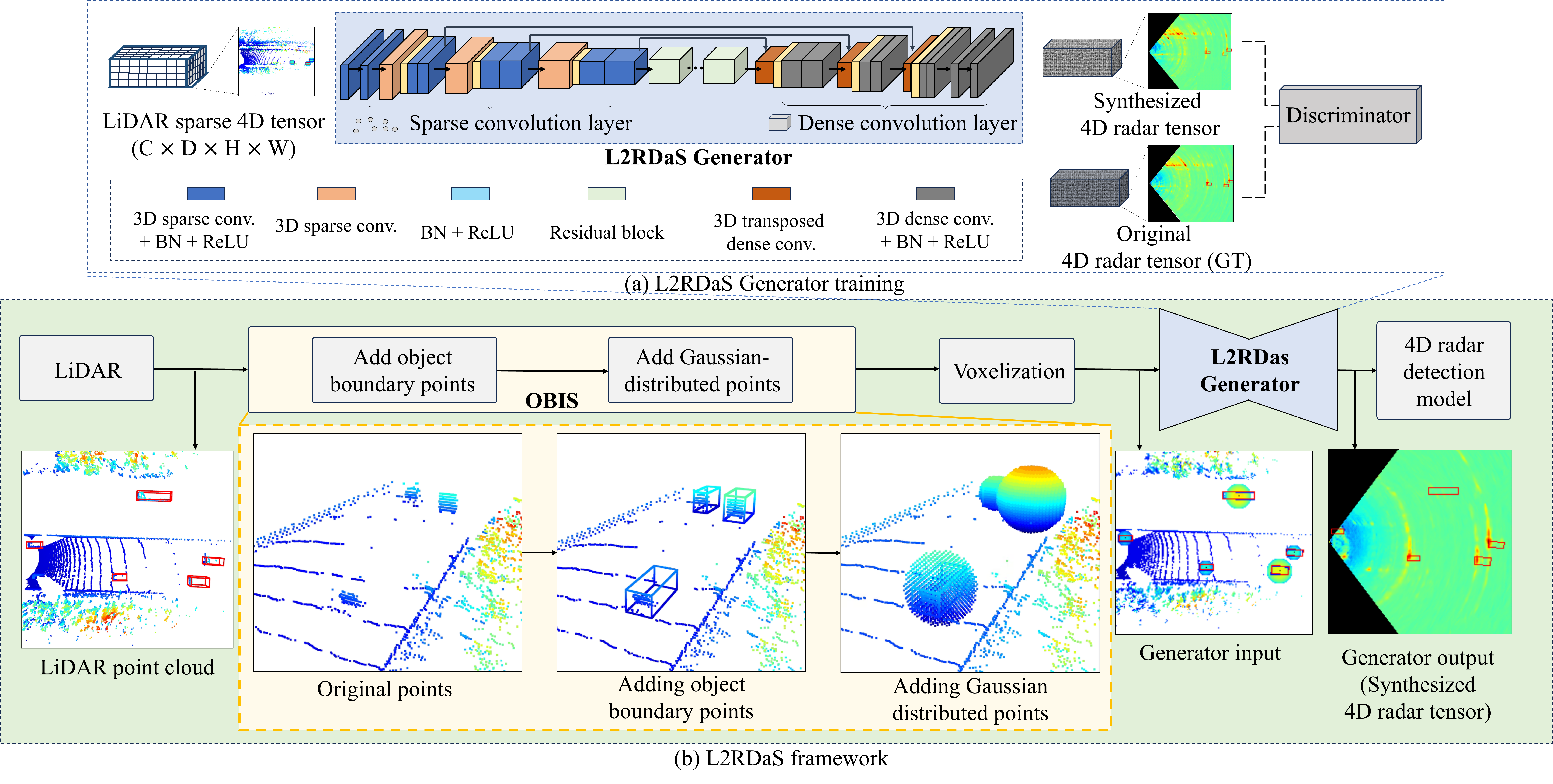} 
    \caption{Overall L2RDaS framework. The core components of the framework are the L2RDaS Generator and the OBIS module.}
    \label{fig:overall_framework}
\end{figure*}

We propose L2RDaS, a framework for synthesizing 4D radar tensors from existing LiDAR point cloud data. The goal is to synthesize spatially informative radar tensors that preserve the full 3D spatial structure—range, azimuth, and elevation—while realistically reproducing radar-specific characteristics such as target reflections and clutter. The overall architecture of the L2RDaS framework is illustrated in Fig.~\ref{fig:overall_framework}.

\subsection{L2RDaS Generator Model} \label{sec:method_generator}
The proposed L2RDaS Generator follows a cGAN \cite{cGAN} architecture based on pix2pixHD \cite{pix2pixhd}, employing an encoder–decoder structure with skip connections to preserve spatial detail, as illustrated in Fig.~\ref{fig:overall_framework}(a). A multi-scale discriminator evaluates the realism of the synthesized 4D radar tensors \cite{pix2pixhd}. This design is widely used in radar synthesis research \cite{l2r, radar_simulate, gan_radar_synth} and serves as a robust baseline.

We adopt a data-driven deep learning framework instead of physics-based simulation, as the former enables flexible training from real sensor data without requiring detailed radar hardware specifications or computationally expensive ray tracing. Accordingly, we selected cGAN as the base model for the L2RDaS Generator. We also explored alternative generative models, including VAEs \cite{VAE} and diffusion models \cite{diffusion}. Prior work \cite{ganvsvae} shows that GANs are more effective at preserving both local and global features than VAEs, due to the presence of a discriminator that discourages over-averaging and helps retain realistic textures \cite{pix2pix}. While diffusion models can produce high-quality samples \cite{ganvsdiffusion}, their slow iterative sampling process (e.g., ~32 seconds per image) limits their practicality for radar tensor synthesis. In contrast, cGAN offer a favorable balance between fidelity and inference speed, making them suitable for synthesizing spatially informative 4D radar tensors.

The L2RDaS Generator takes LiDAR point cloud data as input, offering explicit 3D spatial information consistent with the spatial characteristics of 4D radar tensors. The sparsity of LiDAR data also allows the use of sparse convolution \cite{sparse_conv} operations, improving computational efficiency and scalability for large-scale 3D data. The output is a 4D radar tensors in Cartesian coordinates, where each voxel encodes radar reflection intensity across range, azimuth, and elevation. While original radar tensors are typically represented in polar coordinates, we apply interpolation to convert them into Cartesian coordinates. This Cartesian representation enables precise alignment with LiDAR points and offers a denser, more uniform mapping of distant regions, which are often underrepresented in polar grids. Although Doppler information is not included because single-frame LiDAR lacks the temporal cues required to estimate motion, the synthesized tensors still capture the full 3D spatial distribution of radar reflections. Despite this limitation, synthesizing spatially rich 4D radar tensors remains highly valuable for improving object detection performance, particularly in adverse weather conditions where spatial information is critical.

\subsubsection{Model Architecture}
The L2RDaS Generator is based on the pix2pixHD architecture \cite{pix2pixhd}, which adopts a U-Net \cite{unet} encoder–decoder structure, but it is redesigned to handle 3D spatial information. Before entering the encoder, the LiDAR point cloud is voxelized, following Lee et al. \cite{l2rtranslation_voxel}, to preserve the original spatial structure.

The encoder employs sparse 3D convolution layers \cite{sparse_conv} to efficiently reduce spatial resolution while handling the sparsity of LiDAR input. To prevent feature blurring and strengthen feature extraction, submanifold sparse convolution layers \cite{submanifold} are inserted after each downsampling step. This design allows the network to compress spatial information while preserving sharp feature boundaries and enhancing local detail.

The decoder reconstructs 4D radar tensors using 3D dense convolution layers, enabling complete spatial coverage and producing tensors suitable for downstream tasks. Skip connections between encoder and decoder layers help retain detailed spatial information throughout the network.

The discriminator retains the multi-scale design of pix2pixHD but is adapted for 3D inputs, with each scale using 3D convolutional layers to assess the realism of synthesized 4D radar tensors across multiple resolutions.

\subsubsection{Objective Functions}
L2RDaS adopts an adversarial training framework with a generator \( G \) and three multi-scale discriminators \( D_1, D_2, D_3 \), following the pix2pixHD structure \cite{pix2pixhd, l2r}. The overall objective combines conditional adversarial loss, discriminator feature matching loss, and L1 reconstruction loss:
\begin{equation}
\sum_{k=1}^{3} \left( \mathcal{L}_{cGAN}(G, D_k) + \lambda_{FM} \mathcal{L}_{FM}(G, D_k) \right) + \lambda_{L1} \mathcal{L}_{L1}(G)
\label{eq:full_loss}
\end{equation}
Here, \( \lambda_{FM} \) and \( \lambda_{L1} \) are hyper-parameters that balance the contribution of each term. The definitions of each component are as follows:

\paragraph{Conditional Adversarial Loss}
This loss encourages the L2RDaS Generator to produce 4D radar tensors that are indistinguishable from real data. It is computed for each discriminator \( D_k \) as:
\begin{align}
\mathcal{L}_{cGAN}(G, D_k) = & \; \mathbb{E}_{x, y}[\log D_k(x, y)] \nonumber \\
& + \mathbb{E}_{x}[\log(1 - D_k(x, G(x)))]
\label{eq:cgan_loss}
\end{align}
where \( x \) is the input LiDAR voxel, \( y \) is the real 4D radar tensors, and \( G(x) \) is the synthesized output.

\paragraph{Feature Matching Loss}
To stabilize training and improve realism, we employ a feature matching loss \cite{pix2pixhd}, which minimizes the L1 distance between intermediate features extracted from real and synthesized tensors across discriminator layers:
\begin{equation}
\mathcal{L}_{FM}(G, D_k) = \mathbb{E}_{x,y} \sum_{i=1}^{T} \frac{1}{N_i} \left\| D_k^{(i)}(x, y) - D_k^{(i)}(x, G(x)) \right\|_1
\label{eq:fm_loss}
\end{equation}
where \( D_k^{(i)} \) denotes the output of the \( i \)-th layer of discriminator \( D_k \), and \( T \) is the number of layers used for feature extraction. \( N_i \) represents the number of elements in the \( i \)-th feature map, used for normalization.

\paragraph{L1 Loss}
The L1 loss penalizes voxel-wise differences between the synthesized and real 4D radar tensors, encouraging accurate recovery of reflection intensity values:
\begin{equation}
\mathcal{L}_{L1}(G) = \mathbb{E}_{x, y} \left[ \| G(x) - y \|_1 \right]
\label{eq:l1_loss}
\end{equation}
This loss stabilizes the L2RDaS Generator and complements the adversarial and feature-level losses to enhance reflection fidelity and spatial realism.

\subsection{OBIS Module}
Although we adopt a modified U-Net \cite{unet} architecture to synthesize 4D radar tensors, directly applying the model to datasets lacking native 4D radar tensors and expanding the training data did not yield substantial improvements on their own. We observed that in the synthesized tensors, radar measurements corresponding to objects were slightly shifted compared to real radar tensors. This misalignment stems from the inherently low resolution of radar sensors: unlike LiDAR, which offers fine-grained spatial precision, radar measurements exhibit larger positional deviations. To quantify this, we generated 1000 random objects and measured the average displacement of their center points. In the K-Radar dataset \cite{KRadar}, with LiDAR voxelized at a grid resolution of 0.05 meters, we observed an average center point shift of 0.102 meters. In contrast, radar, using a 0.4-meter voxel resolution, showed an average shift of 1.215 meters—over 10 times larger. Additionally, for small or distant objects, LiDAR measurements often contain fewer than 10 points, which was frequently observed in our experiments. This sparsity imposes a significant limitation when converting to 4D radar tensors, as it often leads to the generation of blurry radar tensors with degraded spatial accuracy.

To address these limitations, we introduce the OBIS module, which enriches LiDAR input with additional object-level cues to enhance the spatial fidelity and reflection realism of the synthesized 4D radar tensors, as illustrated in the yellow box of Fig.~\ref{fig:overall_framework}(b).

\subsubsection{Adding Object Boundary Points}
As noted, the low spatial resolution of radar results in object center shifts exceeding 1 meter. To reduce this error, we add supplementary points along object boundaries, providing the L2RDaS Generator with explicit geometric cues to refine object localization during synthesis. These boundary points also help in better delineating object contours, especially in cluttered environments where objects are densely packed.

The boundary points are generated using 3D bounding box annotations before LiDAR is passed into the L2RDaS Generator. Points are uniformly sampled at 0.1-meter intervals along each bounding box edge. To maintain the overall intensity distribution, each boundary point is assigned the average intensity of the LiDAR frame. A dedicated channel is added to the LiDAR input, marking these as edge points via one-hot encoding.

\subsubsection{Adding Gaussian-Distributed Points}
Sparse LiDAR point cloud leads to incomplete object shapes during radar tensor synthesis, as skip connections alone are insufficient to recover fine spatial structure. When such sparse data are upsampled into radar tensors, the resulting measurement distributions tend to be overly smoothed or blurred.

To compensate for this, we inject auxiliary points arranged spherically around each object center. These act as spatial anchors, enriching the input with denser geometric information to guide the generator in producing spatially informative radar tensors. The added points form a 3D Gaussian distribution based on their distance from the object center. These Gaussian values are encoded into additional semantic channels per object type (e.g., Sedan or Bus/Truck), allowing the network to probabilistically represent each object’s extent and to distinguish between classes by reading object-specific channels. This augmentation mitigates potential misalignments between LiDAR and radar (e.g., calibration or timing errors) and enables the generator to produce radar tensors that better reflect the spatial information characteristic of real-world radar measurements.

\section{Experiments} \label{sec:experiment}
\subsection{Dataset and 4D Radar Detection Models} \label{sec:experiment_dataset_models}
We use the K-Radar dataset \cite{KRadar} to train and evaluate the proposed L2RDaS framework as well as the downstream object detection models. RTNH \cite{KRadar}, the baseline model from the K-Radar benchmark, serves as our primary detector and is designed to operate directly on 4D radar tensors. In addition to RTNH, we evaluate L2RDaS using two lightweight yet effective 4D radar-based object detection models—RadarPillar-Net \cite{radarpillarnet} and RPFA-Net \cite{rpfa_net}—to assess the generalizability of performance gains from data expansion and augmentation.

RadarPillar-Net \cite{radarpillarnet} encodes radar spatial coordinates, Doppler information, and power separately into pillar features, preserving each physical property independently during detection. RPFA-Net \cite{rpfa_net} builds upon this method by incorporating global context into the pillar encoding process to improve heading angle estimation and detection accuracy. While RTNH is a voxel-based detection model, most high-performing radar object detectors are based on the PointPillars \cite{pointpillars} architecture \cite{smurf, radarmfnet}. To reflect this trend and to evaluate the isolated impact of data expansion and GT-Aug, we additionally included these two PointPillars-based models in our experiments due to their simplicity.

The K-Radar dataset provides synchronized 4D radar, high-resolution LiDAR, and RGB camera data, along with ego-motion information (RTK-GPS and IMU), collected under diverse weather conditions (e.g., clear, foggy, snowy) and various road environments (e.g., highways, urban roads, alleyways). For L2RDaS training, we utilize 3D radar tensors (referred to as 3DRT-XYZ), obtained by applying mean pooling along the Doppler axis of the original 4D radar tensors (4DRT) and interpolating into Cartesian space. To reduce computational and memory cost during training, RTNH converts 4D radar tensors into point cloud representations using a percentile-based sparsification method, which selects the top \(k\%\) of radar power values. We apply the same process to the synthesized radar tensors, using a 7\% threshold, to ensure consistency across real and generated data inputs for all detection models.

\subsection{Implementation Details and Training} \label{sec:experiment_training}
The training region of interest (ROI) is defined based on the measurement range of the K-Radar system: \([0, 76.8], [-38.4, 38.4], [-2, 10.8]\) along the \(x\), \(y\), and \(z\)-axes, respectively. Since L2RDaS Generator synthesizes 4D radar tensors from LiDAR, training was performed using data collected under clear weather conditions. The model was trained on an NVIDIA RTX 3090 GPU with a batch size of 1, a learning rate of 0.001, and the Adam optimizer \cite{adam}, for 40 epochs.

We use the following notation throughout the experiments: \(R_{\text{dataset}}^{\text{Real}}\) denotes real 4D radar tensors directly obtained from sensors, while \(R_{\text{dataset}}^{\text{Syn}}\) denotes 4D radar tensors synthesized by the L2RDaS framework from LiDAR input. For instance, \(R_{\text{K-radar}}^{\text{Real}}\) refers to real 4D radar data from the K-Radar dataset, while \(R_{\text{K-radar}}^{\text{Syn}}\) refers to the corresponding synthesized radar tensors. Similarly, \(L_{\text{dataset}}^{\text{Real}}\) denotes the original LiDAR point cloud data, and \(R_{\text{dataset, GT-Aug}}^{\text{Syn}}\) represents radar tensors generated using GT-Aug and the L2RDaS framework.

To evaluate the performance and effectiveness of the L2RDaS framework, we conducted the following experiments:
\begin{enumerate}
    \item \textbf{Synthesis Evaluation:} This experiment was conducted to assess how realistically L2RDaS can synthesize 4D radar tensors that resemble real sensor measurements. We trained the L2RDaS on the K-Radar training set and evaluated the quality of the synthesized radar tensors \(R_{\text{K-radar}}^{\text{Syn}}\) by comparing them with real tensors \(R_{\text{K-radar}}^{\text{Real}}\) from the test set using PSNR and SSIM metrics. Details of the evaluation metrics are provided in Section~\ref{sec:results_metrics}.

    \item \textbf{Detection Performance Evaluation:} Since no prior studies provide a comparable method for synthesizing spatially informative 4D radar tensors, it is difficult to directly evaluate realism using quantitative metrics alone. Therefore, we conducted this experiment to indirectly assess how well the synthesized tensors preserve object-level features. We trained RTNH using the synthesized tensors \(R_{\text{K-radar}}^{\text{Syn}}\) and evaluated detection performance on the real test set to determine whether the generated features are sufficiently informative for 4D radar-based object detection.

    \item \textbf{Dataset Expansion:} This experiment was conducted to verify whether L2RDaS can improve model generalization across diverse driving scenarios by synthesizing 4D radar tensors from external datasets that lack native radar tensors. We synthesized 4D radar tensors from KITTI, nuScenes, VoD, and Dual-Radar datasets—denoted as \(R_{\text{KITTI}}^{\text{Syn}}, R_{\text{nuScenes}}^{\text{Syn}}, R_{\text{VoD}}^{\text{Syn}}, R_{\text{Dual-Radar}}^{\text{Syn}}\)—and incorporated them into the K-Radar training set \(R_{\text{K-radar}}^{\text{Real}}\). We then evaluated performance on the original K-Radar test set to assess whether training with a synthetically expanded dataset yields improved detection performance. \cite{KRadar, kitti, nuscenes, vod, dualradar}

    \item \textbf{4D Radar GT-Aug:} To explore additional applications of L2RDaS, we conducted an experiment using LiDAR-based GT-Aug. Specifically, we applied object-level augmentation to the original LiDAR point clouds in the K-Radar dataset \(L_{\text{K-radar}}^{\text{Real}}\), and then synthesized the corresponding 4D radar tensors \(R_{\text{K-radar, GT-Aug}}^{\text{Syn}}\) using L2RDaS. This experiment was designed to verify whether such augmented tensors could be leveraged to improve detection performance, thereby demonstrating the practical feasibility of 4D radar GT-Aug using L2RDaS.
\end{enumerate}

All detection experiments focus on the 'Sedan' class, following prior K-Radar-based studies \cite{KRadar, rtnh+, echoesfusion, dpft, towardsRadarFusion, l4dr}, which adopted Sedan as the standard target for fair comparison, as it is the most widely distributed class in the dataset.

\subsection{Metric} \label{sec:results_metrics}
To evaluate the quality of the synthesized 4D radar tensors, we used PSNR and SSIM, following the method in \cite{l2r}. Both metrics compare the synthesized tensors with GT tensors and provide complementary insights.

PSNR measures the absolute pixel-wise similarity between the synthesized and GT data. A higher PSNR indicates that the synthesized 4D radar tensors closely match the original in terms of raw power, reflecting how well noise and distortions are suppressed \cite{PSNRvsSSIM}.

SSIM focuses on the structural and perceptual similarity between the two data. It captures how well the synthesized 4D radar tensors preserve important spatial patterns, such as object boundaries and overall scene layout, which is critical for downstream perception tasks \cite{SSIM}.

Although the radar tensors are 3D, both PSNR and SSIM are traditionally applied to 2D images. Therefore, we reduced the tensors to 2D by mean pooling along the height. Additionally, because radar power values span a very large dynamic range (up to \(10^{13}\) in the K-Radar dataset), we applied logarithmic normalization to make the data visually interpretable before evaluation.

For evaluating object detection performance through dataset expansion and GT-Aug, AP based on intersection over union (IoU) was employed. Detections with an IoU of 0.3 or higher were considered as true positives (TP) \cite{kitti_eval}. AP was computed separately for both bird's eye view (BEV) (${{AP}_{BEV}}$) and 3D bounding box predictions (${{AP}_{3D}}$).

\subsection{Experimental Results}
\begin{figure*}[ht]
    \centering
    \includegraphics[width=1\textwidth]{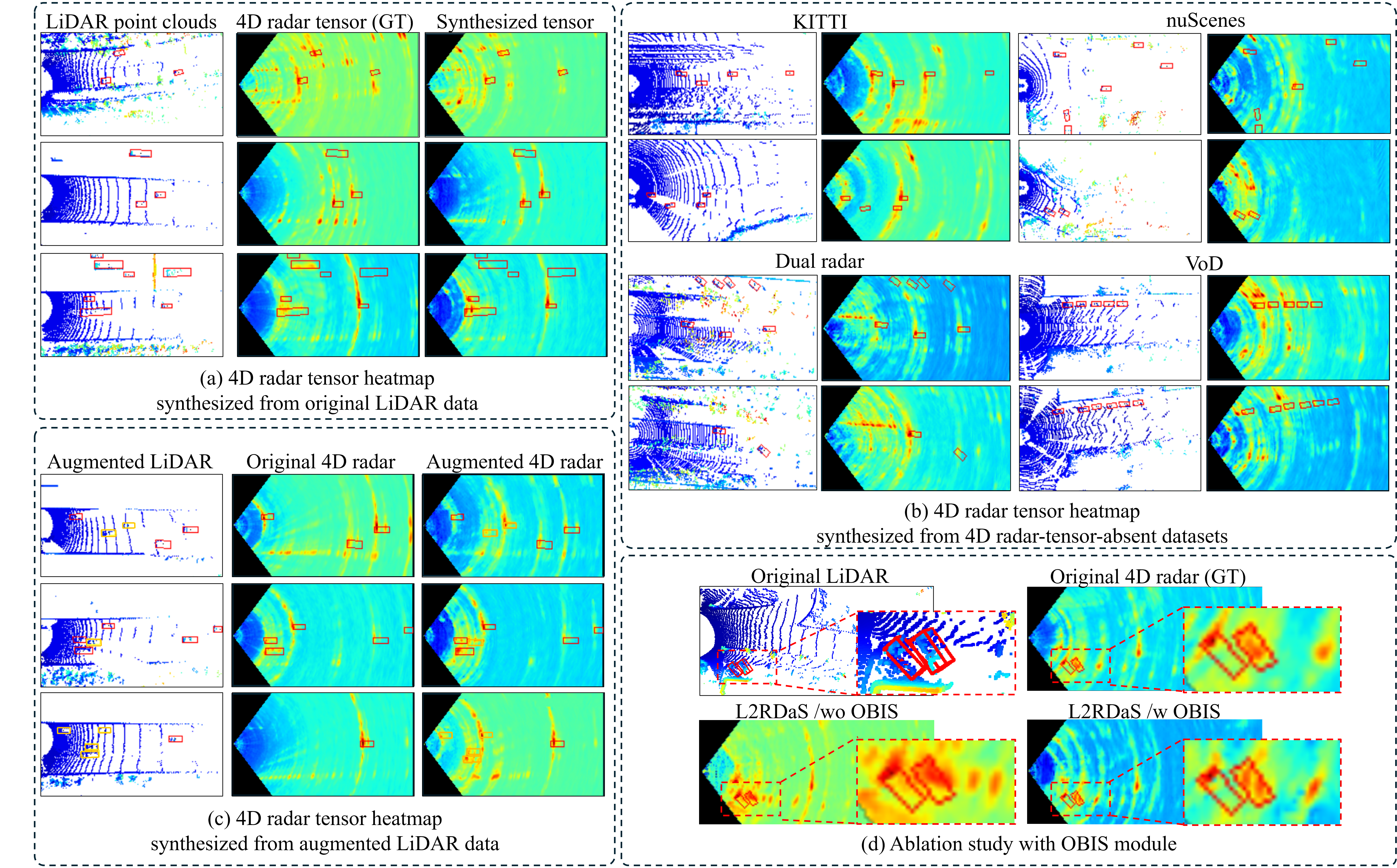} 
    \caption{Qualitative results of the proposed L2RDaS framework. (a) Comparison with GT tensors on the K-Radar test set. (b) Results of applying L2RDaS to existing autonomous driving datasets that do not provide 4D radar tensors. (c) GT-Aug results; orange boxes indicate newly added objects. (d) Effect of the OBIS module in the ablation study.}
    \label{fig:results}
\end{figure*}

This section presents both qualitative and quantitative results for the four experiments described in Section~\ref{sec:experiment_training}. Fig.~\ref{fig:results} visualizes 4D radar tensor heatmaps synthesized by L2RDaS using test samples from K-Radar, KITTI, nuScenes, VoD, and Dual-Radar datasets \cite{KRadar, kitti, nuscenes, vod, dualradar}. The synthesized 4D radar tensors are transformed into BEV representations by applying mean pooling along the height axis. 4D radar power values are logarithmically scaled for better visual contrast, and the color mapping follows the jet colormap, where blue indicates low intensity and red indicates high intensity.

Table~\ref{tab:result_PSNR_SSIM} reports the quantitative similarity between the synthesized and ground-truth 4D radar tensors in Experiment (1), using PSNR and SSIM metrics. Table~\ref{tab:result_obj} summarizes the object detection performance for Experiments (2) to (4), reporting AP metrics to evaluate the effectiveness of the synthesized data for both training and augmentation purposes.

\begin{table}[ht]
\begin{center}
\footnotesize
\caption{Quantitative evaluation of L2RDaS and ablation study on decoder design and OBIS module.}
\label{tab:result_PSNR_SSIM}
\begin{tabular}{c|ll|cc}
\hline \hline
\multirow{2}{*}{Index} & \multicolumn{2}{c|}{Method}                                   & \multicolumn{2}{c}{Metrics} \\ \cline{2-5} 
                       & \multicolumn{1}{c}{Decoder Layer} & \multicolumn{1}{c|}{OBIS} & PSNR (dB) ↑     & SSIM ↑    \\ \hline
(1)                    & Sparse Conv.                      & X                         & 29.551          & 0.828     \\
(2)                    & Dense Conv.                       & X                         & 30.054          & 0.872     \\
(3) L2RDaS             & Dense Conv.                       & O                         & \textbf{31.006}          & \textbf{0.897}     \\ \hline \hline
\end{tabular}%
\end{center}
\end{table}

\begin{table}[ht]
\begin{center}
\caption{Object detection performance using synthesized 4D radar tensors generated by L2RDaS. \(R_{\text{All}}^{\text{Syn}}\) denotes the combined dataset of \(R_{\text{KITTI}}^{\text{Syn}}, R_{\text{nuScenes}}^{\text{Syn}}, R_{\text{Dual-Radar}}^{\text{Syn}},\) and \(R_{\text{VoD}}^{\text{Syn}}\).}
\label{tab:result_obj}
\begin{tabular}{lcl|cc}
\hline \hline
\multicolumn{3}{c|}{Method} &
  \multirow{3}{*}{\shortstack{${{AP}_{BEV}}$\\(\%)}}&
  \multirow{3}{*}{\shortstack{${{AP}_{3D}}$\\(\%)}}\\ \cline{1-3}
\multicolumn{1}{c}{\multirow{2}{*}{\shortstack{Detection\\model}}} &
  \multirow{2}{*}{Experiment} &
  \multicolumn{1}{c|}{\multirow{2}{*}{Train data}} &
   &
   \\
\multicolumn{1}{c}{}             &     & \multicolumn{1}{c|}{}     &                &                \\ \hline
\multirow{4}{*}{RTNH\cite{KRadar}}            & -   & \(R_{\text{Kr}}^{\text{Real}}\)                & 50.32          & 42.82          \\
                                 & (2) & \(R_{\text{Kr}}^{\text{Syn}}\)                & 43.14          & 36.40           \\
                                 & (3) & \(R_{\text{Kr}}^{\text{Real}}\) + \(R_{\text{All}}^{\text{Syn}}\)       & \textbf{53.57} & \textbf{45.92} \\
                                 & (4) & 
                                 \(R_{\text{Kr}}^{\text{Real}}\) + \(R_{\text{Kr, aug}}^{\text{Syn}}\) & 53.11          & 45.79          \\ \hline
\multirow{4}{*}{\shortstack{RadarPillar\\-Net\cite{radarpillarnet}}} & -   & \(R_{\text{Kr}}^{\text{Real}}\)                 & 43.77          & 39.71          \\
                                 & (2) & \(R_{\text{Kr}}^{\text{Syn}}\)                  & 40.22          & 32.73          \\
                                 & (3) & \(R_{\text{Kr}}^{\text{Real}}\) + \(R_{\text{All}}^{\text{Syn}}\)       & 46.35          & \textbf{43.82} \\
                                 & (4) & \(R_{\text{Kr}}^{\text{Real}}\) + \(R_{\text{Kr, aug}}^{\text{Syn}}\) & \textbf{46.46} & 43.63          \\ \hline
\multirow{4}{*}{\shortstack{RPFA\\-Net\cite{rpfa_net}}}            & -   & \(R_{\text{Kr}}^{\text{Real}}\)                & 37.85          & 35.68          \\
                                 & (2) & \(R_{\text{Kr}}^{\text{Syn}}\)                  & 36.26          & 33.77          \\
                                 & (3) & \(R_{\text{Kr}}^{\text{Real}}\) + \(R_{\text{All}}^{\text{Syn}}\)       & \textbf{44.77} & 37.07 \\
                                 & (4) & \(R_{\text{Kr}}^{\text{Real}}\) + \(R_{\text{Kr, aug}}^{\text{Syn}}\) & 43.61          & \textbf{40.89}          \\ \hline \hline
\end{tabular}%
\end{center}
\label{tab:my-table}
\end{table}

\subsubsection{Synthesizing Evaluation}
The qualitative results in Fig.~\ref{fig:results}(a) show that L2RDaS is capable of synthesizing 4D radar tensors that closely resemble the GT tensors, even for unseen inputs. The synthesized outputs successfully capture cluttered regions and preserve realistic spatial distributions of reflection power, as observed in the heatmap patterns.

Quantitative results in Table~\ref{tab:result_PSNR_SSIM} further support these observations. To the best of our knowledge, no prior method has attempted to synthesize spatially informative 4D radar tensors from LiDAR data, making direct comparisons unavailable. Nonetheless, our model achieves an average PSNR of 31.01dB and an SSIM of 0.90, which fall within the range typically considered acceptable for 8-bit data in lossy compression scenarios \cite{psnr_quality}. These findings indicate that the synthesized 4D radar tensors preserve structural fidelity and realistic measurement distributions, which are critical for downstream autonomous driving perception tasks such as object detection.

\subsubsection{Detection Performance Evaluation}
To further validate the utility of the synthesized 4D radar tensors, we trained and evaluated three 4D radar object detection models: RTNH (baseline), RadarPillar-Net, and RPFA-Net, as described in Section~\ref{sec:experiment_dataset_models} \cite{KRadar, radarpillarnet, rpfa_net}.

The RTNH model trained with synthesized data achieved approximately 85\% of the performance compared to the same model trained on real 4D radar tensors. Notably, RadarPillar-Net and RPFA-Net reached up to 90\% performance. These results consistently demonstrate that the synthesized 4D radar tensors are sufficiently informative for object detection, despite minor accuracy drops. However, since the L2RDaS Generator was trained using the same \( R_{\text{K-radar}}^{\text{Real}} \) training set from which the synthesized data \( R_{\text{K-radar}}^{\text{Syn}} \) was derived, the high detection performance may partially result from overfitting to the characteristics of the original dataset. To address this limitation, Experiment (3) investigates whether L2RDaS can generalize to external datasets by training detection models with synthesized 4D radar tensors derived from unseen domains. This enables an indirect evaluation of whether the generated radar tensors exhibit realistic distributions beyond the training set.

\subsubsection{Dataset Expansion for Training}
To evaluate the data expansion capability of L2RDaS, we synthesized 4D radar tensors from KITTI, nuScenes, VoD, and Dual-Radar datasets, as visualized in Fig.~\ref{fig:results}(b). Although these datasets do not natively provide 4D radar tensors, L2RDaS was able to synthesize corresponding tensors from LiDAR inputs.

We then retrained the detection models using an expanded training set that combined the original K-Radar data \( R_{\text{K-radar}}^{\text{Real}} \) with the synthesized tensors \( R_{\text{All}}^{\text{Syn}} \)from the external datasets. For the representative baseline model RTNH, this led to an improvement of 3.25\% in ${{AP}_{BEV}}$ and 3.10\% in ${{AP}_{3D}}$, compared to training on K-Radar alone. On average across all three detection models, L2RDaS-based dataset expansion achieved average increases of 4.25\% in ${{AP}_{BEV}}$ and 2.87\% in ${{AP}_{3D}}$. These results demonstrate the potential of L2RDaS to expand the diversity of training data and improve model generalization across diverse driving scenarios.

\subsubsection{4D Radar GT-Aug}
We evaluated the effectiveness of applying GT-Aug to 4D radar tensors by training the detection model with \( R_{\text{K-radar, GT-Aug}}^{\text{Syn}} \). Fig.~\ref{fig:results}(c) illustrates examples of the augmented 4D radar tensors. In the visualization, red bounding boxes indicate objects originally present in the data, while orange bounding boxes represent objects newly inserted through GT-Aug. These additional objects are seamlessly integrated into the overall measurement distribution, including surrounding clutter. 

In RTNH, training with the augmented 4D radar tensors improved ${{AP}_{BEV}}$ by 2.79\% and ${{AP}_{3D}}$ by 2.97\% compared to training without augmentation. On average across all three detection models, GT-Aug achieved average increases of 3.75\% in ${{AP}_{BEV}}$ and 4.03\% in ${{AP}_{3D}}$. These results demonstrate that L2RDaS can effectively support object-level augmentation in 4D radar tensors, leading to improved detection performance.

\subsection{Ablation Study}
We conducted an ablation study to validate two key design choices of L2RDaS: (1) the use of dense convolution layers in the decoder of the L2RDaS Generator, and (2) the impact of the OBIS module.

\subsubsection{Validation of L2RDaS Generator Decoder Design}
The L2RDaS Generator synthesizes 4D radar tensors from sparse LiDAR point cloud. Its decoder employs 3D dense convolution layers to ensure complete spatial coverage and to produce spatially informative outputs suitable for downstream autonomous driving perception tasks. To evaluate the effectiveness of this design, we compared two variants: one using 3D dense convolution layers and another using 3D sparse convolution layers in the decoder, both excluding the OBIS module.

Quantitative results based on PSNR and SSIM are reported in Table~\ref{tab:result_PSNR_SSIM}. The dense convolution configuration achieved higher PSNR and SSIM scores than its sparse convolution counterpart, with gains of 0.503dB and 0.044, respectively. These findings suggest that dense convolution plays a critical role in generating high-quality 4D radar tensors from sparse LiDAR inputs.

\subsubsection{Validation of OBIS Module}
The OBIS module enriches sparse LiDAR point cloud by injecting object-level cues, including boundary points and Gaussian-distributed points centered around object locations. These enhancements provide additional structural context to the L2RDaS Generator, helping the decoder preserve sharp object measurements during 4D radar tensor synthesis, as illustrated in Fig.~\ref{fig:results}(d).

To assess its contribution, we trained the L2RDaS Generator with and without the OBIS module while keeping the decoder fixed to the dense convolution configuration. As shown in Table~\ref{tab:result_PSNR_SSIM}, the inclusion of OBIS improved PSNR by 0.952dB and SSIM by 0.025, demonstrating its effectiveness in enhancing reflection fidelity and structural consistency of the synthesized 4D radar tensors.

\section{Conclusions} \label{sec:conclusion}
This paper presented L2RDaS, a novel framework for synthesizing spatially informative 4D radar tensors from LiDAR data available in existing autonomous driving datasets. By incorporating a modified U-Net architecture and the OBIS module, the proposed method effectively preserves spatial structure and reflection characteristics, enabling training without additional sensor deployment or data collection. L2RDaS enhances model generalization by expanding datasets with synthetic radar tensors and supports object-level ground-truth augmentation. Future work includes leveraging temporal LiDAR to synthesize 4D radar tensors that include Doppler information and improving alignment with real radar distributions.




\section*{ACKNOWLEDGMENT}

This work was supported by the National Research Foundation of Korea(NRF) grant funded by the Korea government(MSIT) (No. 2021R1A2C3008370).


\bibliographystyle{IEEEtran}
\bibliography{IEEEabrv, references}

\end{document}